\documentclass[10pt,twocolumn,letterpaper]{article}

\usepackage[datasets]{wacv}   
\IfFileExists{axessibility.sty}{\usepackage[accsupp]{axessibility}}{}

\usepackage{enumitem}   
\usepackage{amsmath}
\usepackage{booktabs}   

\definecolor{wacvblue}{rgb}{0.21,0.49,0.74}
\usepackage[pagebackref,breaklinks,colorlinks,allcolors=wacvblue]{hyperref}

\def\wacvPaperID{916}
\def\confName{WACV}
\def\confYear{2027}

\title{Same Outputs, Different Scores:\\
Auditing the Unchecked Graders of Multimodal LLM Benchmarks}

\author{Ahmad Jarjis Hasan\\
Independent Researcher\\
{\tt\small ahmad.hasan2k8@gmail.com}
\and
Jonathan Lwowski\\
Nebius\\
{\tt\small jonathan.lwowski@nebius.com}
}

\newcommand\blfootnote[1]{%
  \begingroup\renewcommand\thefootnote{}\footnote{#1}\addtocounter{footnote}{-1}\endgroup}

\begin{document}
\maketitle

\begin{abstract}
Every multimodal-LLM multiple-choice benchmark ends with a \emph{scoring rule}
that converts a model's response, often free-form text rather than a clean option letter, into a
right/wrong verdict, buried in its code. Unlike the
dataset, this rule is chosen without justification and left unexamined, yet it sets the published
number. We reproduce the official scorers of eight peer-reviewed benchmarks on five open vision-language
models (VLMs) that fit a single 16\,GB GPU, and introduce a blinded human-plus-LLM-judge adjudication
(Cohen's $\kappa=0.956$) that recovers what the model actually answered. The scoring rule alone moves
accuracy by up to \textbf{22 points} on identical outputs, yet \emph{exactly} $0$ wherever a model emits
clean option letters, so it hides on strong-model leaderboards. It is the fallback and not the rambling
that does the damage: the benchmark with the \emph{highest} non-compliance we measure has an artifact of
exactly zero, because its scorer abstains instead of guessing; a reasoning model with non-compliance
$0.000$ still swings $9.5$ points with which regex match of its trace the parser keeps. No standard rule is reliable: snapping a non-answer to
its closest option credits answers never given yet buries correct ones, abstention discards answers
stated in words, and likelihood scoring follows answer-token priors, so a single grade conflates
capability with luck rather than measuring either. We release all code, labels, and the protocol, and argue benchmarks should
report a non-compliance rate and at least two rules, not one unexamined grade.
\end{abstract}

\section{Introduction}
\label{sec:intro}

\begin{figure}[t]
  \centering
  \includegraphics[width=\linewidth]{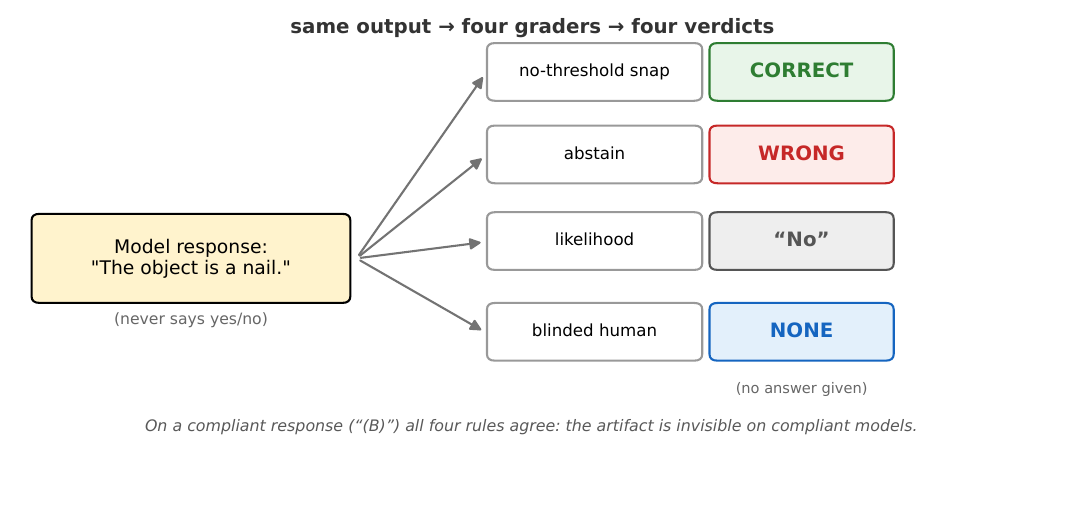}
  \caption{\textbf{The grade comes from the grader, not the model.} A single rambling response receives
  contradictory verdicts under four scoring rules shipped by published benchmarks; none matches what a
  blinded human says the model answered. On a compliant response the rules agree, which is why the
  artifact is invisible on leaderboards of strong, format-following models.}
  \label{fig:hero}
\end{figure}

\blfootnote{\copyright\,2027 IEEE. Personal use of this material is permitted. Permission from IEEE must be obtained for all other uses, in any current or future media, including reprinting/republishing this material for advertising or promotional purposes, creating new collective works, for resale or redistribution to servers or lists, or reuse of any copyrighted component of this work in other works. Accepted at WACV 2027.}
Consider a vision-language model asked whether an industrial part is defective, with options ``(A) Yes''
and ``(B) No.'' It replies: \emph{``The object is a nail.''} The model never answered the question. Yet
MMAD~\cite{mmad}, an ICLR'25 benchmark, scores this response \textbf{correct}: its grader calls
\texttt{difflib.get\_close\_matches(response, options, cutoff=0.0)}, which with a zero threshold snaps
any string to the lexically nearest option: here ``Yes,'' by character overlap, not base rate. An abstaining grader scores
the same response \emph{wrong}. A likelihood grader returns whatever the model's logits prefer. A human
reading it records that the model gave \emph{no answer}. Four graders, four verdicts, one output
(Fig.~\ref{fig:hero}). The grade came from the grader, not the model.

This is not a quirk of one benchmark. Every multimodal-LLM multiple-choice benchmark must decide what to do when a
model's free-form text does not cleanly name an option, and that decision, a few lines at the bottom
of a scoring script, is made differently across the field: a no-threshold snap (MMAD~\cite{mmad},
OmniMedVQA~\cite{omnimedvqa}, PMC-VQA~\cite{pmcvqa}), a random fallback (MMMU~\cite{mmmu},
ScienceQA~\cite{scienceqa}), a default class (POPE~\cite{pope}), abstention, or likelihood scoring. These
benchmarks cleared top-tier peer review and are cited widely, but reviewers judged the \emph{datasets};
the grader was an unexamined implementation detail that produced numbers which looked reasonable, so
it went unopened. This paper opens it.

\begin{figure*}[t]
  \centering
  \includegraphics[width=0.92\linewidth]{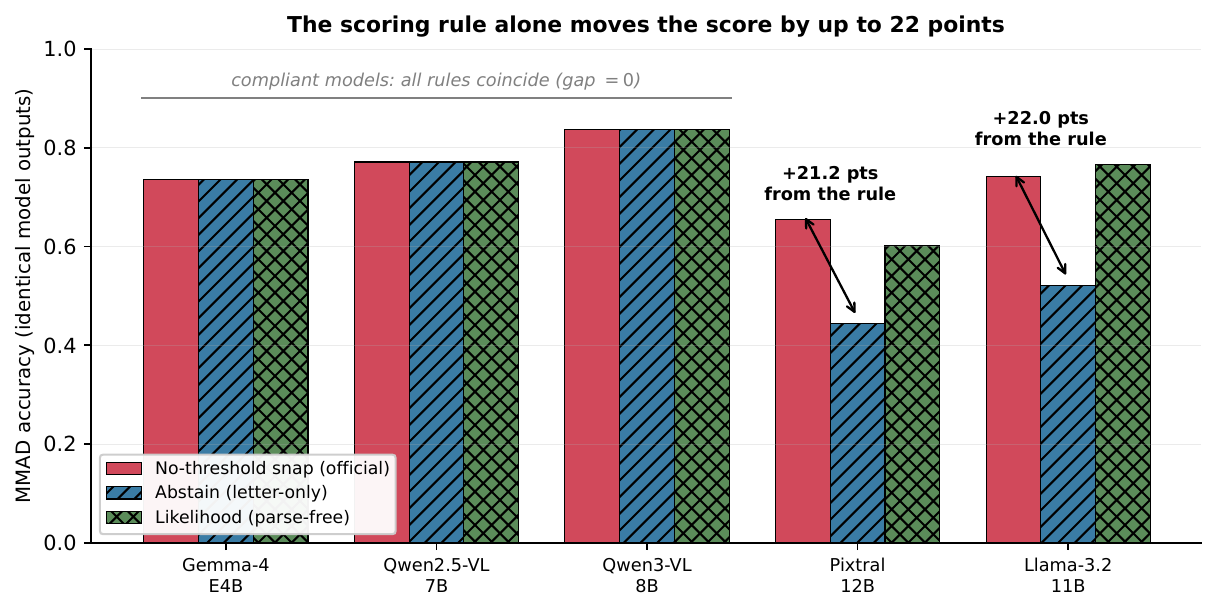}
  \caption{\textbf{The scoring rule alone moves the score by up to 22 points.} MMAD accuracy on
  \emph{identical} stored model outputs under three parse-failure scoring rules. The three
  format-compliant models score the same under every rule (gap $=0$); on the two models that ramble,
  the no-threshold closest-option snap inflates accuracy by $21$--$22$ points over abstention, with the likelihood
  rule different again. Exact on all $1{,}670$ items, deterministic given the outputs.}
  \label{fig:swing}
\end{figure*}

\paragraph{Why the obvious fix is not enough.}
``Just parse the letter and mark the rest wrong'' (abstention) is one choice among several, and it is not
obviously right: a model that answers ``No.'' in words, or reasons correctly and is then cut off before
emitting a letter, is marked wrong. Conversely, the no-threshold snap rewards rambles that never commit.
The rules are not interchangeable, and which one a benchmark ships changes its leaderboard. We scope
this audit to the \emph{locally deployable} regime (open models of ${\le}13$B on one consumer GPU),
the binding constraint for the applications these benchmarks target: privacy-bound medical VQA and
air-gapped industrial inspection, where a frontier API model is not an option. Where a model emits clean letters the rules
coincide, so the choice is invisible there; in the regime practitioners are confined to, the scores
that guide deployment inherit it.

\paragraph{What prior work leaves open.}
A line of work studies the \emph{format sensitivity} of multimodal-LLM scoring: first-token logits versus
decoded text on answers the model \emph{did} give in parseable form~\cite{lyu2024c,robustness}, and
whether models should be allowed to abstain on unanswerable inputs~\cite{upd}. Standardized harnesses do
not dissolve the problem: VLMEvalKit~\cite{vlmevalkit}, which powers the OpenVLM leaderboard, falls back
to \texttt{random.choice} once its extraction retries fail and to a fixed sentinel under its parse-only
policy: the two idioms we audit, shipped as defaults. The closest prior work
anchors answer \emph{extraction} to human judgment for text LLMs and does annotate non-answers,
refusals and truncation among them~\cite[App.~A.3]{molfese2025rightanswer}; what it does not do is
\emph{grade} them, recording a miss instead, and it varies prompt format at a fixed extractor rather
than the scorer at fixed outputs. We audit what the released scorer \emph{credits} when \emph{no}
answer is extractable, a different contrast, and find $21$--$22$ points. The reason ours can be this
large is mechanical. Text LLMs mostly emit a parseable letter; on the open VLMs
practitioners can actually deploy, the fallback fires on $29$--$37\%$ of items. Multimodal is where the
fallback stops being a rounding error and starts setting the number, which is why this audit had to be
done here, in the spirit of evaluation reality-checks in other
subfields~\cite{lucic2018gans,musgrave2020reality}.

\paragraph{Scope of evidence.} That the parse-failure rule \emph{varies and is unreported} across the
field is a code audit of eight peer-reviewed benchmarks (Table~\ref{tab:idioms}), six of which
manufacture a verdict and two of which do not. That \emph{no rule reliably recovers what the model
answered} is a stronger claim, demonstrated where we have human ground truth: binary MMAD and four-option
medical PMC-VQA. We anchor it on binary MMAD by design, not by limitation: yes/no is where blinded
readers agree almost perfectly on what a model said (inter-annotator $\kappa=0.91$), so it is where the
grader's fabrication is \emph{certifiable} rather than asserted. Breadth is carried by the field-wide
survey, proof by the binary anchor.

\paragraph{Contributions.}
\begin{itemize}[leftmargin=1.2em,itemsep=1.5pt,topsep=2pt]
  \item \textbf{An audit of released graders} (Sec.~\ref{sec:graders},~\ref{sec:results}). We reproduce
  the official scoring code of eight peer-reviewed benchmarks on five open 16\,GB-class VLMs; the
  parse-failure rule alone shifts accuracy by up to \textbf{22 points} (Fig.~\ref{fig:swing}) on
  identical outputs and exactly $0$ on format-compliant models; the artifact is invisible until a
  model rambles.
  \item \textbf{A human-certified adjudication protocol} (Sec.~\ref{sec:setup},~\ref{sec:results}).
  A blinded human plus LLM-judge procedure recovers what the model actually answered ($\kappa=0.956$,
  released), and shows \emph{no} standard rule is reliable, in both directions: the snap credits answers
  never given on one model yet \emph{deflates} another's correct answers, abstention discards answers
  given, and likelihood scoring is biased by answer-token priors.
  \item \textbf{A practice recommendation and release} (Sec.~\ref{sec:reco}). We release all
  code, labels, and the protocol,\footnote{\url{https://github.com/ahmadhasan2k8/grader-audit}} and recommend benchmarks report a
  non-compliance rate and at least two scoring rules rather than one unexamined grade.
\end{itemize}

\section{Related Work}
\label{sec:related}

\paragraph{Format sensitivity of multimodal-LLM scoring.}
\citet{lyu2024c} report large disagreements between a model's first-token logit and its decoded
free-text answer on multiple-choice questions; others study option-order and prompt robustness and the
generative-versus-discriminative gap~\cite{robustness}; \citet{rosenthal2025unexplored} further show that
semantically-neutral prompt-format changes alone reorder MLLM leaderboards, as does prompt format in text
LLMs~\cite{sclar2024formatspread}; \citet{alzahrani2024benchmarks} perturb the benchmark and the
answer-selection method together and move leaderboard rank by up to eight positions. All vary the
\emph{input}, or analyze answers produced in \emph{parseable} form. We hold the outputs byte-identical and
vary only the parse-\emph{failure} rule, anchoring each to an external human label rather than to another
automated rule.

\paragraph{Answer extraction and standardized evaluation.}
A parallel line improves how a \emph{parseable} answer is recovered or standardized: trained extractors
that pinpoint the option a model chose~\cite{xfinder}, standardized prompt-and-parse protocols~\cite{olmes},
reproducibility practices for evaluation harnesses~\cite{lmeval}, and, for text LLMs, evidence that
answer-extraction choices disagree with human judgment on MCQA~\cite{molfese2025rightanswer}, and that
the extraction algorithm alone shifts reasoning-model results~\cite{jo2025finding}; in the
multimodal setting, MMBench~\cite{liu2024mmbench} adds an LLM choice-extractor and circular evaluation
inside a new benchmark. A stronger response is to abandon the format outright:
\citet{chandak2025matching} show that free-form \emph{answer matching}, graded against a reference by a
capable model, reaches human-level agreement where multiple choice does not, and argue evaluation should
move off discriminative MCQ. That is a prescription for the benchmarks of the future; ours is a diagnosis
of the ones in use today. These make the parser better, the protocol uniform, the benchmark harder, or the
format different, but none audits, for existing multimodal benchmarks, what the scorer does when the
parser \emph{fails}. We target exactly that residual: the parse-failure fallback, measured against what
the model actually answered.

\paragraph{Abstention and unanswerable inputs.}
A separate line argues multimodal LLMs should be allowed to abstain and that benchmarks mismeasure when they force
a choice~\cite{upd}. We share the premise but apply it to the \emph{grader}: the parse-failure fallback
is a forced choice imposed by the grading code, independent of whether the model tried to abstain.

\paragraph{Evaluation reality-checks.}
A recurring genre re-examines a trusted evaluation practice and finds it unreliable: GAN metrics under
equal compute budgets~\cite{lucic2018gans}, deep metric-learning gains under fair comparison
\cite{musgrave2020reality}. The genre is now active for LLM and multimodal evaluation and well
represented at datasets-and-benchmarks venues: judge reliability~\cite{zheng2023mtbench}, benchmark
robustness and format noise~\cite{wang2024mmlupro}, multimodal harness reality-checks~\cite{lmmseval},
lexical matching that underestimates generative models against human
judgment~\cite{kamalloo2023evaluating}, grading logic that over-credits agents~\cite{agenticbench2025}, and,
closest to us in modality, dataset validity for LVLMs~\cite{chen2024mmstar} (visual necessity and
leakage). We are the reality-check for the \emph{scoring rule} of multimodal-LLM multiple-choice
evaluation. Unlike a general reproduction, we hold each benchmark's own official parser fixed and change
only the parse-failure fallback, which attributes the measured gap to the scoring rule alone.

\paragraph{Benchmarks studied.}
We audit MMAD~\cite{mmad}, MMMU~\cite{mmmu}, ScienceQA~\cite{scienceqa}, OmniMedVQA~\cite{omnimedvqa},
PMC-VQA~\cite{pmcvqa}, and POPE~\cite{pope}, and use MathVista~\cite{mathvista} and
MathVision~\cite{mathvision} as the cases where an extraction step and an abstaining rule, respectively,
defuse the same idiom. All are peer-reviewed.

\section{The Graders}
\label{sec:graders}

Let a multiple-choice item have $k$ options and gold answer $g$; the model emits free text $r$. A scoring
rule is a function $s(r)\to\{1,\dots,k\}\cup\{\bot\}$, where $\bot$ means abstain (scored wrong). The rule
has a \emph{parser} that tries to extract a clean option from $r$ and a \emph{fallback} for when the
parser fails. Both embed unreported decisions: where the parser looks and how it breaks ties among
several candidate letters, and what the scorer does when it finds nothing. Our headline measurements
isolate the fallback, which is where these benchmarks differ most and which is almost never justified or
reported; Sec.~\ref{sec:notopt} shows the parser side is not inert either, and that it is the side the
field's move to reasoning models activates. Table~\ref{tab:idioms} lists the released idiom for each benchmark we audit; it is a \emph{code} audit verified at pinned commits, and we quantify the effect (Sec.~\ref{sec:results}) for the five benchmarks whose official prompt yields a clean reproduction, treating ScienceQA, whose released regex accepts exactly one answer format that our models rarely produce (Sec.~\ref{sec:notopt}), as a code-presence finding only.

\begin{table*}[t]
\centering\small
\begin{tabular}{lll}
\toprule
Benchmark & Venue & Parse-failure fallback (released code) \\
\midrule
MMAD~\cite{mmad}        & ICLR'25      & no-threshold snap (\texttt{get\_close\_matches},\,cutoff $0$) \\
OmniMedVQA~\cite{omnimedvqa} & CVPR'24 & no-threshold snap (\texttt{SequenceMatcher}) \\
PMC-VQA~\cite{pmcvqa}   & Comm.\,Med.'24 & no-threshold snap (\texttt{SequenceMatcher}) \\
MMMU~\cite{mmmu}        & CVPR'24      & \texttt{random.choice} over options \\
ScienceQA~\cite{scienceqa} & NeurIPS'22 & \texttt{random.choice} over options \\
POPE~\cite{pope}        & EMNLP'23     & default class (``yes'') \\
\midrule
MathVista~\cite{mathvista} & ICLR'24   & snap, \emph{preceded by} an LLM extraction step \\
MathVision~\cite{mathvision} & NeurIPS'24 D\&B & abstain: no fallback, unparseable scored wrong \\
\bottomrule
\end{tabular}
\caption{\textbf{The parse-failure fallback is an unreported design choice that varies across
peer-reviewed benchmarks.} We read each released scorer to recover the rule; none of these papers
reports the accuracy its fallback is responsible for, which is the quantity we measure. The bottom two rows are the counterexamples that make the choice visible: an extraction buffer
and an abstaining rule each defuse the idiom (Sec.~\ref{sec:notopt}).}
\label{tab:idioms}
\end{table*}

\paragraph{Three families.}
\textbf{No-threshold snap} maps $r$ to the lexically closest option with no similarity cutoff, so any
string yields an option; it is \emph{directional}, concentrating verdicts on whichever option text is the
common shape of an answer (e.g.\ ``Yes'' in anomaly detection). \textbf{Random fallback} returns a
uniform option, undirected, $\approx 1/k$ correct per firing. \textbf{Default class} returns a fixed
label. Against these, two rules do not manufacture a verdict from an unparseable answer:
\textbf{abstention} (score wrong) and \textbf{likelihood} (never parse text; pick the option the model
assigns highest likelihood; unconditional, never gated by parsing, specified in the supplement). A natural reading is that the first three are ``bad'' and the last two are
``safe.'' Section~\ref{sec:results} shows the truth is messier: abstention and likelihood each fail in
their own way, and which rule is least wrong is model-dependent.

\paragraph{The field already edits these rules silently.} Two of the released scorers we audit
have changed a scoring decision without reporting it as one. MMAD's thirteen released harnesses share one
answer parser but index its output inconsistently: eleven keep the \emph{last} matching letter, two keep
the \emph{first}, a difference worth nothing on the models it was built for and $9.5$ accuracy points on
a reasoning model (Sec.~\ref{sec:notopt}). MMMU parses the entire response; its
successor MMMU-Pro~\cite{mmmupro} restricts the search to the text after the last \texttt{``Answer:''}
(\texttt{evaluate.py}, whose docstring cites the very MMMU file it diverges from), a change announced
in neither paper. Scoring rules are revised in practice, just not in print.

\section{Setup}
\label{sec:setup}

\paragraph{Models and reproduction.}
We use five current open vision-language models that each run 4-bit on a single 16\,GB GPU:
Gemma-4-E4B~\cite{gemma4}, Qwen2.5-VL-7B~\cite{qwen25vl}, Qwen3-VL-8B~\cite{qwen3vl},
Pixtral-12B~\cite{pixtral}, and Llama-3.2-11B-Vision~\cite{llama3}. For each benchmark we clone the
official repository at a pinned commit and call its released scoring function on our models' outputs; we
do not reimplement the scorer. Non-compliance is not an artifact of under-specified prompting: MMAD's
released prompt already ends ``Answer with the option's letter from the given choices directly,'' and
Pixtral still fails to emit a letter on $489/1{,}670$ items ($29\%$), once replying ``Here's the option's
letter from the choices directly.'' We generate every response with greedy decoding (temperature $0$),
so the stored outputs are deterministic and carry no sampling variance; this is what licenses our
``same outputs, every rule'' design, since all rule differences are then attributable to the scorer
alone. Models are loaded in 4-bit (bitsandbytes NF4, \texttt{bfloat16} compute) with a generation budget
of $16$ tokens on the letter-answer benchmarks and up to $768$ on the chain-of-thought ones. ScienceQA is
the one benchmark whose non-compliance is a \emph{format} mismatch rather than a failure to answer
(Sec.~\ref{sec:notopt}), so we treat its magnitude as a code-presence
finding only (Sec.~\ref{sec:graders}). Elsewhere non-compliance is a property of the models, not the budget:
quadrupling it to $64$ tokens (Tab.~\ref{tab:budget}) leaves MMAD non-compliance nearly flat and the snap
artifact large, because these models describe the image or answer in words rather than emit a letter. As
a validity check we score our stored outputs twice, once with MMAD's \emph{own} released harness
(\texttt{caculate\_accuracy\_mmad}) and once with our wrapper, and their per-item verdicts agree to
four decimals on four of five models (its printed headline further macro-averages the normal/anomaly
split; we report plain accuracy). The fifth, Pixtral, differs by $0.2$ points because the harness keeps
the \emph{last} regex match where our wrapper keeps the first: immaterial here, since only Pixtral ever
emits two option letters ($19$ of $8{,}350$ responses) and the artifact is identical under both
conventions, but the same unreported choice that decides $9.5$ points on a reasoning model
(Sec.~\ref{sec:notopt}).

\begin{table}[t]
\centering
\caption{\textbf{The artifact is robust to the generation budget.} Quadrupling the MMAD budget from $16$
to $64$ tokens leaves both models' non-compliance above $0.26$ and the snap$-$abstain artifact above $17$
points; the rambling is a model behavior, not a truncation artifact. Full $1{,}670$-item set.}
\label{tab:budget}
\small
\begin{tabular}{lcccc}
\toprule
& \multicolumn{2}{c}{Non-compliance} & \multicolumn{2}{c}{Snap$-$abstain} \\
\cmidrule(lr){2-3}\cmidrule(lr){4-5}
Model & $16$ & $64$ & $16$ & $64$ \\
\midrule
Pixtral-12B   & $0.29$ & $0.27$ & $+0.212$ & $+0.183$ \\
Llama-3.2-11B & $0.37$ & $0.33$ & $+0.220$ & $+0.171$ \\
\bottomrule
\end{tabular}
\end{table}

\paragraph{The adjudication protocol.}
To anchor the rules to ground truth we adjudicate what the model actually answered. For each item we
build a \emph{blinded} record containing only the question, the options, and the model's full response;
the gold answer and every rule's verdict are withheld. A reader assigns one label: a letter (the model
committed to that option, by letter or by stating its text), \textsc{none} (it never committed: it
described, hedged, refused, or ran out), or \textsc{multiple} (contradictory). The join from saved
responses to questions is verified by matching the gold sequence (it aborts on mismatch).

The $130$ MMAD items we adjudicate comprise $110$ non-compliant cases (where the rules disagree) and $20$
compliant controls ($55$ and $10$ per model), drawn from the two models that ramble (Pixtral, Llama). \textbf{Two} human annotators,
one author and one annotator naive to the hypothesis, label them independently; on a $40$-item overlap
they agree on $38$ ($95\%$, inter-human Cohen's $\kappa_{\mathrm{HH}}=0.91$), establishing that the
letter/\textsc{none}/\textsc{multiple} rubric is well-posed and not an artifact of one reader. A blinded LLM judge (Claude Opus~4.8, temperature $0$, same rubric) labels the same
items so it can scale beyond the human subset, and agrees with the human label on $127$ of $130$
($97.7\%$, \emph{judge-vs-human} Cohen's $\kappa=0.956$); the three disagreements are boundary cases (a
hedge, a contradictory echo, a buried commitment). We release the manifest, both label sets, the rubric,
and the code; the judge is a frontier model used only as a research-time labeling aid on this released
data, not a deployment-time dependency, and anchoring the decomposition to the human rather than the
judge leaves it unchanged (Sec.~\ref{sec:notopt}). A rule's accuracy is then decomposed against the human label into \emph{over-credit} (it
scores correct on an item the model did not answer with the gold option) and \emph{under-credit} (it
scores wrong on an item the model did answer correctly). We extend the same protocol to a second domain,
medical PMC-VQA ($k{\approx}4$, Sec.~\ref{sec:notopt}); Table~\ref{tab:agreement} summarizes agreement
across both.

\begin{table}[t]
\centering
\caption{\textbf{Adjudication agreement.} The LLM judge tracks the human anchor, and the rubric is
stable across two independent humans. On medical PMC-VQA the raw $\kappa$ is a notch lower, but almost
every disagreement is a \textsc{none}-vs-\textsc{multiple} boundary call on a ramble; collapsing those
two non-committal labels into the one distinction the decomposition uses raises $\kappa$ to $0.95$.}
\label{tab:agreement}
\small
\begin{tabular}{llccc}
\toprule
Comparison & Benchmark & $k$ & $n$ & Cohen's $\kappa$ \\
\midrule
Judge vs.\ human & MMAD    & 2 & 130 & 0.956 \\
Human vs.\ human & MMAD    & 2 & 40  & 0.910 \\
Judge vs.\ human & PMC-VQA & 4 & 60  & 0.839 \\
\bottomrule
\end{tabular}
\end{table}

\section{Results}
\label{sec:results}

\subsection{The scoring rule alone moves the score}
On identical model outputs, swapping only the parse-failure fallback changes accuracy by up to
\textbf{22 percentage points} (Fig.~\ref{fig:swing}); every ``point'' in this paper is a percentage point
of accuracy. On MMAD the no-threshold snap scores Pixtral-12B at $0.656$ and Llama-3.2-11B at
$0.741$, while abstaining scores them at $0.444$ and $0.522$, a $+0.212$ (Wilson 95\% CI
$[0.193,0.232]$) and $+0.220$ (95\% CI $[0.201,0.240]$) gap on the full $n{=}1{,}670$ set. On the
three models that emit clean letters the gap is \emph{exactly} $0$: the artifact is invisible until a
model rambles, and it rides on each paper's specific models' compliance. The effect is smaller but
present elsewhere (MMMU $+0.04$ on the model that rambles most, Gemma, and $+0.09$ at its largest,
Llama's; OmniMedVQA up to $+0.08$;
POPE $\approx\!+0.01$, an easy task where almost everything is compliant); Table~\ref{tab:grid} gives
every benchmark--model cell we measure. The magnitude is set by the
fallback \emph{type} and the option count, not by task difficulty: at matched non-compliance (MMAD
Pixtral $0.29$, MMMU Gemma $0.27$) the swing is $3$--$5\times$ larger on the \emph{easier} binary task
($+0.212$ vs $+0.040$--$0.071$), because a directional two-option snap concentrates on the high-base-rate option
while a $k$-way random fallback disperses at ${\approx}1/k$, so the biggest artifact lands on the
simplest benchmark. A $3$-point shift looks minor
beside MMAD's $22$, but on a competitive benchmark it is often the entire gap between the top model and
the fifth-ranked one, so an unreported fallback can silently invalidate a comparative claim, and the
cost falls on the specialist models these benchmarks exist to serve.

\subsection{No rule recovers ground truth}
Anchored to the human label ($\kappa=0.956$, Sec.~\ref{sec:setup}), each rule fails differently. On the
$110$ non-compliant MMAD items, the model \emph{committed} to an option only $23\%$ of the time, yet the
snap scored $72\%$ ``correct'': it credits answers the model never gave. Abstention has the opposite
failure: on the $17\%$ of those items where the model \emph{did} state the gold answer in words (``No.''), the
letter-only rule marks it wrong. Likelihood scoring fails a third way (Sec.~\ref{sec:notopt}). There is
no neutral default: the snap over-credits, abstention under-credits, and the gap between them on MMAD
($0.21$) is a gap between two flawed rules, not a measurement of the model. The $22$-point figure needs
no adjudication: it is exact on all $1{,}670$ items; adjudication supports only the stronger claim that
neither rule recovers what the model said. The two rules also \emph{span} the
scores a reporting policy can produce on the full set: $[0.444,0.656]$ for Pixtral, $[0.522,0.741]$ for
Llama. We report that as a policy-sensitivity interval, not a bound on adjudicated accuracy: the snap
under-credits even on MMAD (Table~\ref{tab:decomp}), the lower end assumes a parsed letter \emph{is}
the committed answer, which Sec.~\ref{sec:notopt}(v) shows can fail, and on PMC-VQA the ordering
inverts.
A gallery of real responses with each rule's verdict and the blinded human label is in the supplement.

\begin{table*}[t]
\centering
\caption{\textbf{The parse-failure artifact across every benchmark and model we audit.} Each cell is the
accuracy swing attributable to the parse-failure rule alone, with the rate at which that rule fires
(non-compliance) in parentheses, computed on \emph{identical} stored outputs under each benchmark's own
released scorer. ``Non-compliance'' is in each case the rate at which that benchmark's own scorer reaches
its fallback. Because each cell isolates the fallback, no artifact can exceed its firing rate, and none
does. MMMU and ScienceQA cells are one seeded draw of their own \texttt{random.choice} fallback, whose
expectation reorders MMMU's models (supplement); MMAD's snap is deterministic. Two regimes appear. \textbf{Firing:} the fallback manufactures a verdict and moves the
score. \textbf{Defused:} an
extraction buffer (MathVista) or an abstaining rule (MathVision) means the scorer never fabricates.
MathVision is the control that separates prevalence from harm: it has the \emph{highest} non-compliance
of any benchmark here ($0.54$--$0.69$) and an artifact of \emph{exactly} zero, because its scorer
abstains instead of guessing; zero artifact shows the fallback adds no credit, not that extraction is
faithful. Rambling is not the bug; the fallback is.}
\label{tab:grid}
\footnotesize\setlength{\tabcolsep}{4pt}
\begin{tabular}{llccccc}
\toprule
Benchmark & Regime & Gemma-4-E4B & Qwen2.5-VL-7B & Qwen3-VL-8B & Pixtral-12B & Llama-3.2-11B \\
\midrule
MMAD              & firing  & $+0.000$ (0.00) & $+0.000$ (0.00) & $+0.000$ (0.00) & $+0.212$ (0.29) & $+0.220$ (0.37) \\
MMMU              & firing  & $+0.040$ (0.27) & $+0.007$ (0.04) & $+0.047$ (0.17) & $+0.027$ (0.16) & $+0.087$ (0.21) \\
OmniMedVQA        & firing  & $+0.005$ (0.01) & $+0.013$ (0.015) & $+0.010$ (0.01) & $+0.083$ (0.39) & $+0.018$ (0.06) \\
PMC-VQA$^{\ddagger}$ & firing & $+0.015$ (0.04) & $+0.000$ (0.00) & $+0.002$ (0.01) & $+0.097$ (0.61) & $+0.020$ (0.12) \\
POPE              & firing  & $+0.017$ (0.05) & $+0.000$ (0.00) & $+0.003$ (0.003) & $+0.025$ (0.05) & $+0.003$ (0.003) \\
ScienceQA$^{\dagger}$ & firing & $+0.062$ (0.20) & $+0.070$ (0.23) & $+0.027$ (0.07) & $+0.118$ (0.34) & $+0.303$ (0.84) \\
\addlinespace
MathVista         & defused & $+0.004$ (0.03) & $+0.011$ (0.03) & $+0.004$ (0.01) & $+0.017$ (0.04) & $+0.007$ (0.04) \\
MathVision        & defused & $+0.000$ (0.61) & $+0.000$ (0.56) & $+0.000$ (0.61) & $+0.000$ (0.69) & $+0.000$ (0.54) \\
\bottomrule
\end{tabular}\\[2pt]
{\footnotesize $^{\dagger}$ScienceQA's released regex requires the literal string \texttt{The answer is
X.}; the models overwhelmingly answer in a near-miss format instead (Sec.~\ref{sec:notopt}), so the
magnitude is exact on these outputs but the \emph{firing rate} would differ under a different prompt.
\quad $^{\ddagger}$PMC-VQA's matcher is unconditional: besides the fallback tabulated here, it also
\emph{overrides} answers its own parser reads cleanly, a second and larger failure on one model
(Sec.~\ref{sec:notopt}).}
\end{table*}

\begin{figure}[t]
  \centering
  \includegraphics[width=\linewidth]{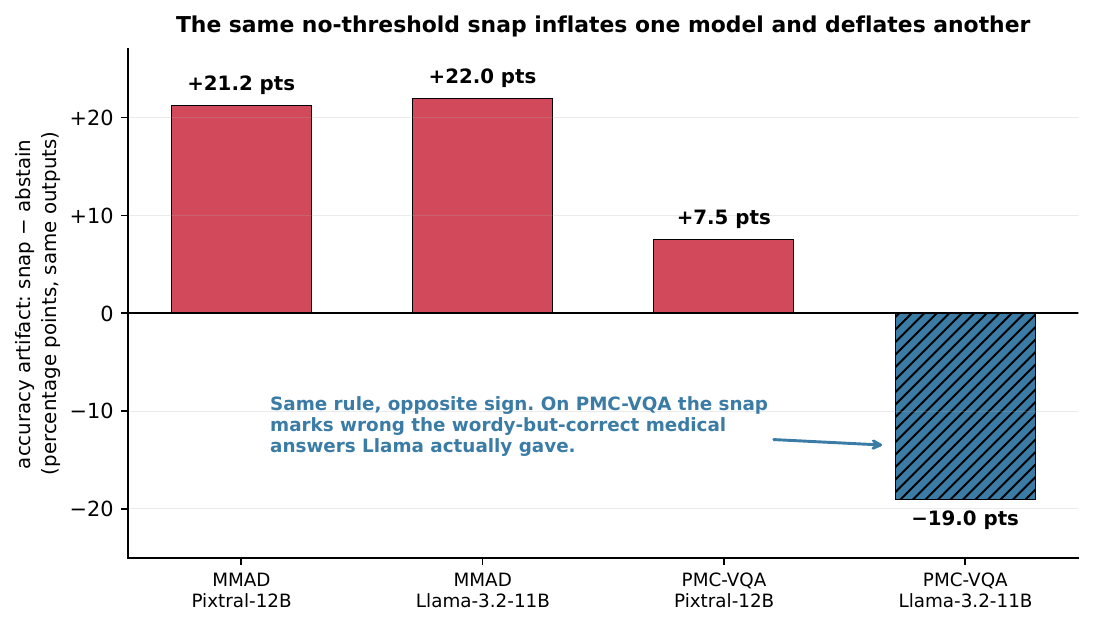}
  \caption{\textbf{The official scorer adds $22$ points to one model and removes $19$ from another.} Signed accuracy swing
  (official scorer $-$ clean parse) on our two human-anchored benchmarks. On MMAD this is
  \emph{entirely} the parse-failure fallback: the two rules agree on every compliant item, so
  $+0.212$/$+0.220$ is attributable to the fallback alone. On PMC-VQA the matcher is unconditional and the
  total decomposes into a fallback term ($+0.097$ Pixtral, $+0.020$ Llama) and an override term
  ($-0.022$, $-0.210$); Llama's $-0.190$ is dominated by the latter (Sec.~\ref{sec:notopt}). Per-item
  mechanism in Tab.~\ref{tab:decomp}.}
  \label{fig:bidir}
\end{figure}

\begin{table}[t]
\centering
\caption{\textbf{The snap's error is bidirectional and model-dependent.} Human-anchored over- vs.\
under-credit on adjudicated items (MMAD anchored to the human label, $n{=}65$; medical PMC-VQA to the
certified judge, $n{=}48$). ``Gave gold'' counts adjudicated items committed to the gold option. On binary MMAD the snap over-credits heavily; on medical PMC-VQA
the same rule \emph{under-credits} Llama on $15/48$ items, $11$ of them items where the fallback itself fired.
Wilson 95\% CIs: MMAD Pixtral over-credit $62\%\,[49,72]$, PMC-VQA Llama under-credit $31\%\,[20,45]$.
Within PMC-VQA Llama the direction itself is significant ($15$ under-credits against $4$ over-credits,
exact binomial $p{=}0.019$), so the sign flip is not a small-sample artifact.}
\label{tab:decomp}
\footnotesize\setlength{\tabcolsep}{3.5pt}
\begin{tabular}{llcccc}
\toprule
Benchmark & Model & $n$ & Gave gold & Over & Under \\
\midrule
MMAD    & Pixtral-12B   & 65 & 11 & 40 (62\%) & 0 (0\%) \\
MMAD    & Llama-3.2-11B & 65 & 23 & 25 (38\%) & 4 (6\%) \\
PMC-VQA & Pixtral-12B   & 48 & 4  & 5 (10\%)  & 2 (4\%) \\
PMC-VQA & Llama-3.2-11B & 48 & 17 & 4 (8\%)   & 15 (31\%) \\
\bottomrule
\end{tabular}
\end{table}

\subsection{The error has no safe direction, and compliance does not fix it}
\label{sec:notopt}
Five further results show no rule is a safe default.
\textbf{(i) The same scorer adds $22$ points to one model and removes $19$ from another} (Fig.~\ref{fig:bidir})\textbf{.} PMC-VQA's matcher is
\emph{unconditional} (it runs on every response), so besides filling parse failures it can override an
answer the parser already read. Separating the two on identical outputs, the fallback alone adds $+0.097$
to Pixtral and $+0.020$ to Llama, while the \emph{override} subtracts $0.210$ from Llama, mapping onto a
different option the answer its own parser reads as \emph{gold} on $153/600$ items ($25.5\%$ of the
benchmark). The aggregate $-0.190$ is the sum, and the larger term is not the fallback, a distinction
our other benchmarks never force, because their matchers run only when the parser fails. Anchored to the
human label the fallback is bidirectional too. A $60$-item audit (judge--human $\kappa=0.84$, $0.95$ on the
commit-vs-no-commit split this turns on) confirms a real per-item effect: the snap \emph{under-credits}
Llama on $31\%$ of items, marking wrong medical answers it actually gave, while over-crediting Pixtral on
$10\%$; the split is unchanged whether unaudited items anchor to the human or the judge ($0.29{\to}0.31$). ``Unreliable,'' not
``optimistic.'' Because the sign of the error flips with the model's verbosity and the dataset's option
format, the grader acts as a \emph{confound}: it injects variance tied to response length rather than
capability, so one unexamined grade measures neither cleanly. A weak probe agrees: on items where a human found no single
committed option, the model's own option likelihood prefers gold at chance for Pixtral ($26/52$)
but above chance for Llama ($24/34$, two-sided $p{=}0.024$), so even the \emph{character} of the snap's unearned
credit is model-dependent; the single-token logit is itself token-prior biased~\cite{holtzman2021surface,wang2024myanswer} (ii), so we do not read
this as a capability claim.
\textbf{(ii) The principled rule can be the most wrong: $0.403$ against the snap's $0.565$.} Parse-free likelihood is not a clean ground
truth: on OmniMedVQA, Llama's option-letter logit collapses onto the token ``A'' on $459/600$ items while
gold is ``A'' only $29\%$ of the time, giving accuracy $0.403$, \emph{below} the snap's $0.565$. Any
single rule, including the principled one, can be the most wrong.
\textbf{(iii) An extraction buffer defuses the idiom: $+10$ points collapses to $+0.9$.} MathVista runs an LLM extraction step before the same
no-threshold snap; scoring the snap on raw responses yields a spurious $+10$-point artifact, but the
faithful pipeline (extraction first, reproduced with a local model) collapses it to $+0.9$ points,
evidence that the fix is a competent parse, not a different fallback.
\textbf{(iv) One character of unreported regex is worth $29.5$ points.} ScienceQA is the purest case, and not for the reason we
first assumed: its released regex demands the literal \texttt{The answer is X.}, and the models answer
\texttt{The answer is (X).} instead. Of Llama's $507$ parse failures, $460$ contain that
parenthesised form and \emph{none} is truncated (median response $110$ characters). The scorer therefore
discards answers that are present and correct, and randomly fabricates in their place: accepting both
spellings (abstaining on the residual failures) raises Llama from $0.448$ to $0.743$ on identical
stored outputs (${\approx}0.80$ keeping that fallback instead). A single character of
unreported regex is worth $29.5$ points to one model, which is the thesis of this paper in its least
ambiguous form.
\textbf{(v) Non-compliance $0.000$, and the grade still moves $9.5$ points.} Every scorer we audit applies its
parser to the \emph{entire} response: harmless when models emitted two characters, not once they
reason first. On MMAD, Qwen3-VL-8B-Thinking is fully format-compliant: non-compliance $0.000$, the
parse-failure fallback fires \emph{zero} times, the artifact this paper measures is exactly $0$, and it
closes its \texttt{</think>} block within budget on $99.8\%$ of $1{,}670$ items. Yet its trace restates
MMAD's option list before answering, so \emph{which} regex match the scorer keeps now decides the grade.
MMAD's eleven majority harnesses keep the last and recover the model's stated answer on
$1666/1666$ items; the two that keep the first disagree with it on $15.9\%$ and score
identical outputs at $0.689$ rather than $0.784$, a $9.5$-point gap (category-clustered $95\%$ CI
$[6.4,13.1]$, $15$ MVTec~\cite{mvtecad} categories). The benchmark ships both numbers and reports neither choice. This
is not a capability claim: correctly scored, the reasoning model ($0.784$) still trails its instruction-tuned
sibling ($0.837$) on these items. It is the sharper form of our recommendation: a non-compliance rate of
$0$ certifies that nothing was fabricated, not that the grade is safe, a pattern
\citet{jo2025finding} document for text LLMs and we find in shipped multimodal scoring code. MathVision~\cite{mathvision} makes
the same point from the opposite direction, abstaining rather than snapping at the highest
non-compliance we measure (Tab.~\ref{tab:grid}): what converts rambling into a corrupted score is the
fallback, not the difficulty.

\paragraph{What we do not claim.}
Holding each benchmark's parser fixed and varying only the fallback, we find no robust leaderboard
reordering: the strongest model stays top-ranked, and a top-1 change on MMMU under abstention is not
significant at our sample size ($p\!=\!0.58$). Where the released \emph{parser} is itself the variable the
picture differs: accepting ScienceQA's near-miss format moves Llama from last to fourth, which is one
reason we report ScienceQA as a code-presence finding rather than folding it into the headline. The
contribution is the unreliability of the grade, not a corrected ranking: stable ranks are the weaker
guarantee, and the same ordering sits on a $21$-point spread in the number practitioners deploy by and
later papers must match.
Nor does a shared harness dissolve it: VLMEvalKit ships the same random fallback (Sec.~\ref{sec:intro}),
and the thirteen harnesses inside MMAD alone disagree with each other by $9.5$ points (item v).

\section{Recommendation}
\label{sec:reco}

No single scoring rule reflects what a model knew or said, so a benchmark should not hide its choice of
rule inside a scoring script and report one number. We recommend three low-cost practices.

\begin{enumerate}[leftmargin=1.4em,itemsep=2pt,topsep=2pt]
  \item \textbf{Report the non-compliance rate} (the fraction of outputs on which the parse-failure
  fallback fires) alongside accuracy. It is one number, and it is the single statistic that exposes how
  much of a score the grader manufactured; when it is large, the headline accuracy is not comparable
  across papers. It is necessary but not sufficient, and reasoning models are where the difference
  starts to matter: a model can be perfectly compliant (non-compliance $0$, the fallback never
  firing) while its grade still turns on where in the response the parser looks
  (Sec.~\ref{sec:notopt}). Report the rate, and state the parser's scope and tie-break alongside it.
  \item \textbf{Report at least two rules.} Publish accuracy under both an abstaining rule (crediting
  only parseable letters) and the closest-option snap, rather than a single grade under either alone.
  On MMAD, where adjudication shows the snap rarely under-credits, the spread approximates the range of
  defensible scores; on PMC-VQA it inverts, the snap scoring Llama \emph{below} the abstaining rule. Unconditionally, the spread is the rule-sensitivity a single
  number hides, and a spread that inverts is itself the warning. Likelihood scoring is a useful parse-free cross-check, not a safe
  default: it too can be the most-wrong rule (Sec.~\ref{sec:notopt}).
  \item \textbf{Adjudicate when it matters.} Our blinded human-plus-LLM-judge protocol ($\kappa=0.956$)
  recovers what the model actually answered for roughly one person-day per benchmark, and the certified
  judge can then label the full non-compliant set automatically, at no further human cost. We release the
  \textbf{adjudication corpus}: one stored generation per item for all eight benchmarks and five models,
  $130$ blinded MMAD adjudications ($40$ dual-annotated by a reader naive to the hypothesis,
  $\kappa_{\mathrm{HH}}=0.91$), $60$ blinded PMC-VQA adjudications, and the certified judge's labels on the
  audited items, with the manifest builder, blinded MCQ tool, judge, and analysis code, under
  permissive licenses (MIT / CC-BY-4.0), with a datasheet and one-command containerized reproduction, so
  any benchmark author can audit what their grader credits.
\end{enumerate}

MathVista already shows the remedy works in the wild: an LLM extraction step placed before the snap
collapses a spurious $+10$-point artifact to $+0.9$ (Sec.~\ref{sec:notopt}). Our protocol operationalizes
and certifies that step rather than inventing one.

None of these is novel machinery: abstention, likelihood, and LLM-as-judge are all in use. The
contribution is the evidence that the choice between them is consequential and currently unexamined, and
a certified, released procedure to make it.

\section{Limitations}
\label{sec:limitations}
We state the boundaries plainly. (i) The human-anchored ground truth is strongest on \emph{binary} MMAD, the easiest case for both the
artifact and the adjudication. We extend the certified judge to medical PMC-VQA ($k{\approx}4$), where a
$60$-item audit gives judge--human $\kappa=0.84$ overall and $0.95$ on the commit-vs-no-commit
distinction the decomposition uses (residual disagreement is almost entirely
\textsc{none}-vs-\textsc{multiple} on rambles, treated identically by the decomposition), so the
cross-benchmark claim does not rest on a single anchor. (ii) Outside MMAD and code-presence-only ScienceQA, the certified fallback magnitudes are typically
$1$--$5$ points, reaching $8$--$10$ on the models that ramble; PMC-VQA's override term reaches $-21$ and
ScienceQA's parser effect $29.5$; the contribution is the unreliability of the grade, not its size. (iii) Varying only the fallback, we find no robust ranking change; varying the
released \emph{parser} does move ranks (Sec.~\ref{sec:notopt}), reported but not built on. (iv) The single-token option logit is a weak
probe of ``latent knowledge''; we use it only to show the snap's credit is model-dependent, and report
where it is itself confounded (Sec.~\ref{sec:notopt}). (v) We test five
open $7$--$12$B models 4-bit on a $16$\,GB GPU; we measure neither proprietary nor much larger models; an 8-bit control
does not restore compliance (supplement), and the artifact is a property of the scorer
\emph{given} the outputs. Compliance is not an
escape route: Sec.~\ref{sec:notopt}(v) reports a model with non-compliance $0.000$ whose grade still
moves $9.5$ points, so the failure does not retire as models strengthen: it migrates from the fallback
to the parser.

\section{Conclusion}
\label{sec:conclusion}
A benchmark score is read as a measurement of the model. But when a model rambles, as the locally
deployable models these benchmarks target routinely do, part of that score is manufactured by a grading
choice no paper wrote down. Scores are therefore not comparable across papers that chose differently,
and the remedy is nearly free: state the rule, report how often it fires, publish more than one number. We give a certified procedure for checking what a grader credits, and hope
the next benchmark affords the last lines of its scoring code the scrutiny it affords its data: the
dataset is not the only thing worth reviewing.

{
    \small
    \bibliographystyle{ieeenat_fullname}
    \bibliography{main}
}

\end{document}